# A 58.6mW Real-Time Programmable Object Detector with Multi-Scale Multi-Object Support Using Deformable Parts Model on 1920x1080 Video at 30fps

Amr Suleiman, Zhengdong Zhang, Vivienne Sze

Massachusetts Institute of Technology, MA, USA

## Abstract

This paper presents a programmable, energy-efficient and real-time object detection accelerator using deformable parts models (DPM), with 2x higher accuracy than traditional rigid body models. With 8 deformable parts detection, three methods are used to address the high computational complexity: classification pruning for 33x fewer parts classification, vector quantization for 15x memory size reduction, and feature basis projection for 2x reduction of the cost of each classification. The chip is implemented in 65nm CMOS technology, and can process HD (1920x1080) images at 30fps without any off-chip storage while consuming only 58.6mW (0.94nJ/pixel, 1168 GOPS/W). The chip has two classification engines to simultaneously detect two different classes of objects. With a tested high throughput of 60fps, the classification engines can be time multiplexed to detect even more than two object classes. It is energy scalable by changing the pruning factor or disabling the parts classification.

## Introduction

Object detection is critical to many embedded applications that require low power and real-time processing. For example, low latency and HD images are important for autonomous control to react quickly to fast approaching objects, while low energy consumption is essential due to battery and heat limitations. Object detection involves not only classification/recognition, but also localization, which is achieved by sliding a window of a pre-trained model over an image. For multi-scale detection, the window slides over an image pyramid (multiple downscaled copies of the image). Multi-scale detection is very challenging as the image pyramid results in a data expansion, which can be more than a 100x in HD images. The high computational complexity of object detection processing necessitates fast hardware implementations [1] to enable real-time processing.

This paper presents a complete object detection accelerator using DPM [2] with a root and 8 parts model as shown in Fig. 1. DPM results in double the detection accuracy compared to rigid template (root only) detection. The 8 parts account for deformation such that a single model can detect objects at different poses (Fig. 6) and increase detection confidence. However, this accuracy comes with a classification overhead of 35x more multiplications (i.e. DPM classification consumes 80% of a single detector power), making multi-object detection a challenge. A software-based DPM object detector is described in [3], which enables detection for 500x500 images at 30fps but requires a powerful fully loaded Xeon 6-core processor and 32GB of memory. In this work, the classification overhead is significantly reduced by two main techniques:

- Classification pruning with vector quantization (VQ) for selective part processing.
- Feature basis projection for sparse multiplications.

## Architecture Overview

Fig. 2 shows the block diagram of our detector architecture, including histogram of oriented gradients (HOG) feature pyramid generation unit and support vector machine (SVM) classification engines. A feature pyramid size of 12 scales (4 octaves, 3 scales/octave) is selected as a trade-off between detection accuracy and computation complexity. The pyramid contains 87K feature vectors, which is 2.7x more features than a typical HD image. To meet the throughput, three parallel histogram and normalize blocks generate the pyramid. Two classification engines share the generated feature to detect two different classes of objects simultaneously. The root and the parts SVM weights can be programmed with a maximum template size of 128x128 pixels. This large size gives the detector the flexibility to detect many objects classes with different aspect ratios. Each SVM engine contains a root classifier for root detection, a pruning block to select candidate roots, and 8 part processing engines for parts detection. Local feature storage in each part engine allows parallelism, reduces the feature storage read bandwidth and enables 7x speedup. Finally, the Deform block uses a coarse-to-fine technique for 2.2x speedup in finding the maximum score in a 5x5 search region for each part after adding the deformation cost.

## Classification Pruning and Vector Quantization

With more than 2.6 million features generated per second, on-the-fly processing is used for root classification similar to [3] for minimal on-chip storage, where partial dot products are accumulated in SRAMs. Using the same approach with parts classification would require large accumulation SRAM sizes (more than 800KB for one classification engine). However, it was observed that if the root score is too low, then the likelihood of detecting an object based on parts is also low. Since parts classification requires significant additional computation, we choose to prune the parts classification when the root score is below a programmable threshold. By pruning 97% of the parts classification (i.e. a 33x reduction of the root candidates that are processed), we achieve a 10x reduction in classification power with negligible 0.03% reduction in accuracy.

To avoid re-computation, HOG features are stored in line buffers to be reused by the part processing engines after pruning (Fig. 3). VQ is used to reduce the feature line buffers write bandwidth (from 44.4MB/s to 2.5MB/s), making its size suitable for on-chip SRAM (from 572KB to 32KB) and eliminating any off-chip storage. Three parallel VQ engines are used to meet the throughput. A programmable 256 clusters centers are stored in a shared SRAM to minimize the read bandwidth. The 143-bit HOG feature vector (13-D, 11-bit each) is quantized to 8 bits per vector, giving a 15x reduction in the overall feature storage size. De-quantization is just a memory read from the feature SRAM.

## Feature Basis Projection

To further reduce the cost of each classifier, the features are projected into a new space where the classification SVM weights are sparse. Zeros multiplications are skipped and only the non-zero weights are stored on-chip. Fig. 4 shows that the percentage of zero weights is increased from 7% to 56% after projection. The programmable basis vectors are designed such that at least 7 out of the 13 dimensions in the weights are zeros. An overhead of a 13-bit flag is stored to label the zeros positions, resulting in a total reduction of the SVM weight SRAM size and read bandwidth by 34%. The basis projection reduces the number of multiplications by 2x and reduces the overall classification power by 43%.

## Evaluation

The accuracy of our accelerator is analyzed on PASCAL VOC 2007 [5], which is a widely used image dataset containing 20 different object classes (aeroplane, bicycle, bird, etc.) in 9,963 images. With 97% pruning, VQ and feature basis projection, 10x fewer classification multiplications and a 3.6x smaller memory size are achieved, leading to 5x reduction in the total power consumption with a drop in the detection accuracy by only 4.8%.

**Implementation and Testing**

The chip is implemented in 65nm CMOS technology. It is tested to process HD images at 30fps in real-time operating at 62.5MHz and 0.77V while consuming 58.6mW, resulting in a peak performance of 1168 GOPS/W and an energy efficiency of 0.94nJ/pixel. Fig. 5 shows the die photo and the chip specifications, along with a sweep for different throughput and energy consumption. At 1.11V, the chip can process HD images up to 60fps while consuming 216.5mW. The power breakdown for different tested configurations is shown in Fig. 6. The DPM classification power is significantly reduced down to only 15% of the total power of a single detector. With features sharing, detecting an additional object class with DPM increases the total power consumption by only 19%. Fig. 6 also shows the chip output with multi-object detection (cars and pedestrians). Comparing to the detector accelerator in [6], our chip boosts the detection accuracy with multi-scale and detecting 8 deformable parts per object while consuming 30% less energy per pixel.

**Acknowledgement**

The authors would like to thank TSMC University Shuttle Program for the chip fabrication and DARPA and TI for funding.

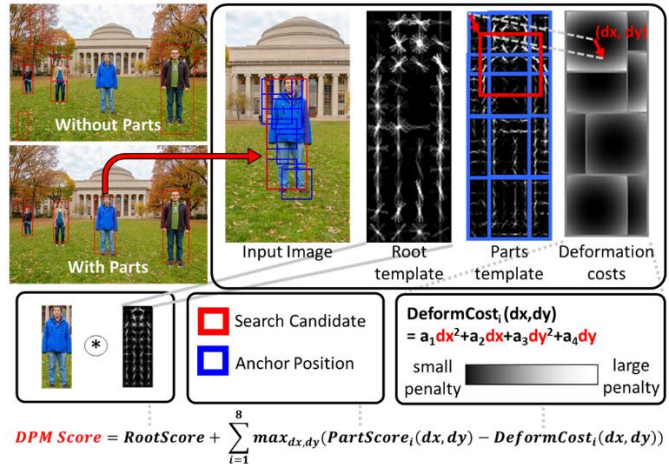

Fig. 1 Detection example with DPM templates and score calculation

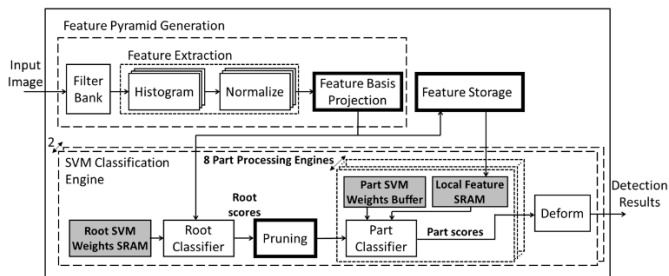

Fig. 2 Block diagram of the DPM object detection accelerator

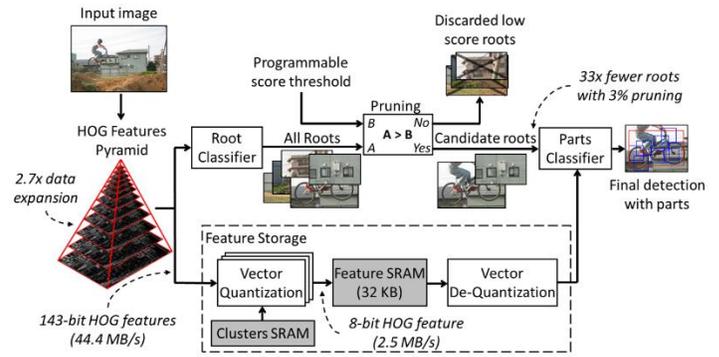

Fig. 3 DPM detection with pruning and vector quantization

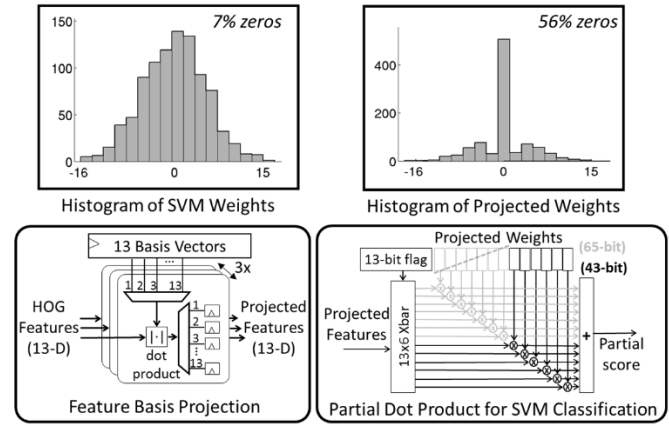

Fig. 4 Feature basis projection for sparse classification

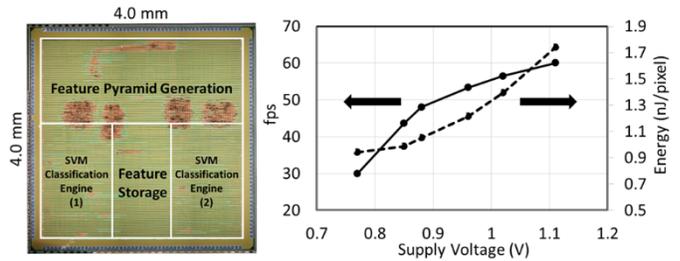

| Technology | 65nm CMOS | Frame rate | 30 – 60 fps |
|---|---|---|---|
| Chip size | 4.0 x 4.0 mm² | Resolution | 1920x1080 |
| Core size | 3.58 x 3.58 mm² | Power | 58.6 – 216.5 mW |
| Logic gates | 3283 kgates | Energy/pixel | 0.94 – 1.74 nJ |
| SRAM | 280.1 KB | GOPS | 68 - 137 |
| Supply | 0.77 – 1.11 V | GOPS/W | 1168.7 – 623.8 |
| Frequency | 62.5 – 125 MHz | GOPS/mm² | 4.25 – 8.56 |

Fig. 5 Die photo and summary of the chip specifications. Numbers are measured with the two detectors running and 97% pruning set.

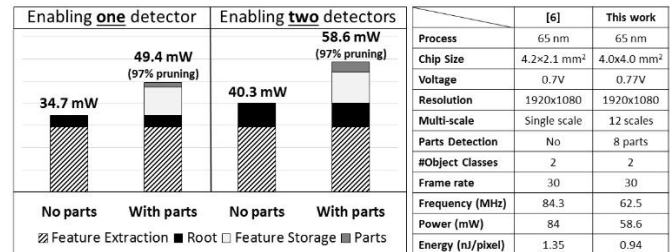

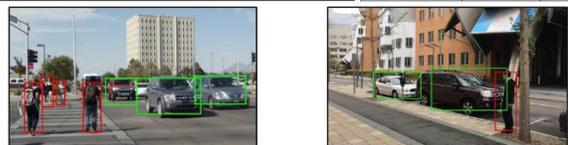

| | [6] | This work |
|---|---|---|
| Process | 65 nm | 65 nm |
| Chip Size | 4.2×2.1 mm² | 4.0x4.0 mm² |
| Voltage | 0.7V | 0.77V |
| Resolution | 1920x1080 | 1920x1080 |
| Multi-scale | Single scale | 12 scales |
| Parts Detection | No | 8 parts |
| #Object Classes | 2 | 2 |
| Frame rate | 30 | 30 |
| Frequency (MHz) | 84.3 | 62.5 |
| Power (mW) | 84 | 58.6 |
| Energy (nJ/pixel) | 1.35 | 0.94 |

Fig. 6 Performance comparisons and detection examples.